\documentclass[runningheads]{llncs}
\usepackage{graphicx}

\usepackage{bbding}
\usepackage{tikz}
\usepackage{comment}
\usepackage{amsmath,amssymb} 
\usepackage{color}
\usepackage{graphicx}
\usepackage{comment}
\usepackage{amsmath,amssymb} 
\usepackage{color}
\usepackage{amsmath}
\usepackage{amsmath,amssymb}
\usepackage{multirow}
\usepackage{array}
\usepackage{wrapfig}

\usepackage[T1]{fontenc}
\usepackage[utf8]{inputenc}
\usepackage{babel}
\usepackage[font=small,labelfont=bf]{caption}

\usepackage{makecell}
\usepackage[pagebackref,breaklinks,colorlinks]{hyperref}
\usepackage{hyperref}
\hypersetup{
    colorlinks=true,
    linkcolor=blue,
    filecolor=blue,      
    urlcolor=blue,
    citecolor=cyan,
}

\usepackage[accsupp]{axessibility}  

\emergencystretch=\maxdimen
\begin{document}
\pagestyle{headings}
\mainmatter
\def\ECCVSubNumber{1347}  

\title{Detecting and Recovering \\ Sequential DeepFake Manipulation} 

\titlerunning{SeqDeepFake}
%
\author{Rui Shao \and
Tianxing Wu \and
Ziwei Liu\thanks{Corresponding author}}
\authorrunning{R. Shao et al.}
%
\institute{S-Lab, Nanyang Technological University \\
\email{\{rui.shao, twu012, ziwei.liu\}@ntu.edu.sg}
\texttt{\url{https://rshaojimmy.github.io/Projects/SeqDeepFake}}}

\maketitle

\begin{abstract}
Since photorealistic faces can be readily generated by facial manipulation technologies nowadays, potential malicious abuse of these technologies has drawn great concerns. Numerous deepfake detection methods are thus proposed. However, existing methods only focus on detecting \textit{one-step} facial manipulation. As the emergence of easy-accessible facial editing applications, people can easily manipulate facial components using \textit{multi-step} operations in a sequential manner. This new threat requires us to detect a sequence of facial manipulations, which is vital for both detecting deepfake media and recovering original faces afterwards. Motivated by this observation, we emphasize the need and propose a novel research problem called Detecting Sequential DeepFake Manipulation (\textbf{Seq-DeepFake}). Unlike the existing deepfake detection task only demanding a binary label prediction, detecting Seq-DeepFake manipulation requires correctly predicting a sequential vector of facial manipulation operations. To support a large-scale investigation, we construct the first Seq-DeepFake dataset, where face images are manipulated sequentially with corresponding annotations of sequential facial manipulation vectors. Based on this new dataset, we cast detecting Seq-DeepFake manipulation as a specific image-to-sequence (\textit{e.g.} image captioning) task and propose a concise yet effective Seq-DeepFake Transformer (\textbf{SeqFakeFormer}). Moreover, we build a comprehensive benchmark and set up rigorous evaluation protocols and metrics for this new research problem. Extensive experiments demonstrate the effectiveness of SeqFakeFormer. Several valuable observations are also revealed to facilitate future research in broader deepfake detection problems. 



\end{abstract}

\section{Introduction}

\begin{figure}[t] 
	\begin{center}
		\includegraphics[height=4cm, width=0.85\linewidth]{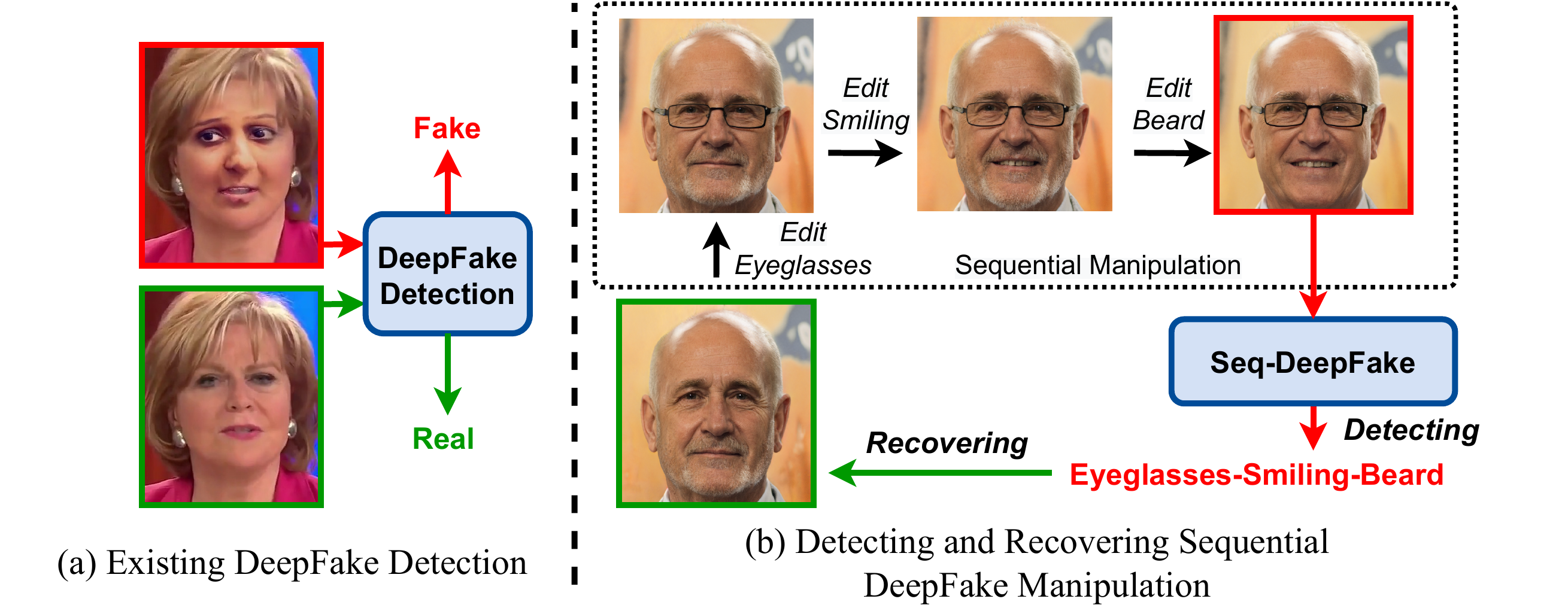}
	\end{center}
	\vspace*{-7mm}
	\caption{Comparison between (a) existing deepfake detection and (b) proposed detecting and recovering sequential deepfake manipulation.}
	\label{fig:deepfakecmp}
	\vspace{-16pt}
\end{figure}

In recent years, hyper-realistic face images can be generated by deep generative models which are visually extremely indistinguishable from real images. Meanwhile, the significant improvement for image synthesis brings security concerns on potential malicious abuse of these techniques that produce misinformation and fabrication, which is known as \textit{deepfake}. To address this security issue, various deepfake detection methods have been proposed to detect such forged faces. As illustrated in Fig.~\ref{fig:deepfakecmp} (a), given the manipulated face image generated by face swap algorithm~\cite{rossler2019faceforensics++} and the original face image, the existing deepfake detection task requires the model to predict the correct binary labels (Real/Fake).

With the increasing popularity of easy-accessible facial editing applications, such as YouCam Makeup\footnote{\tiny{https://apps.apple.com/us/app/youcam-makeup-selfie-editor/id863844475}}, FaceTune2\footnote{\tiny{https://apps.apple.com/us/app/facetune2-editor-by-lightricks/id1149994032}}, and YouCam Perfect\footnote{\tiny{https://apps.apple.com/us/app/youcam-perfect-photo-editor/id768469908}}, it is convenient for people to edit face images in daily life. Compared to existing deepfake techniques mainly carrying out \textit{one-step} facial manipulation~\cite{rossler2019faceforensics++,he2021forgerynet}, we can now easily manipulate face images using \textit{multi-step} operations in a \textit{sequential} manner. As shown in Fig.~\ref{fig:deepfakecmp} (b), the original image can be manipulated by adding eyeglasses, making a bigger smile and removing beard sequentially. This expands the scope of existing deepfake problem by adding sequential manipulation information and poses a new challenge for current \textit{one-step} deepfake detection methods. This observation motivates us to introduce a new research problem --- Detecting Sequential Deepfake Manipulation (\textbf{Seq-Deepfake}). We summarize several key differences between detecting Seq-Deepfake and the existing deepfake detection: 1) rather than only predicting binary labels (Real/Fake), detecting Seq-Deepfake aims to detect sequences of facial manipulations with diverse sequence lengths. For example, the model is required to predict a 3-length sequence as `Eyeglasses-Smiling-Beard' for the manipulated image as shown in Fig.~\ref{fig:deepfakecmp} (b). 2) As illustrated in Fig.~\ref{fig:deepfakecmp} (b), beyond pure forgery detection, we can further \textbf{recover} the original faces (refer to Section 5.4 of Experiments) based on the detected sequences of facial manipulation in Seq-Deepfake. This greatly enriches the benefits of detecting Seq-Deepfake manipulation.

\begin{figure}[t] 
	\begin{center}
		\includegraphics[height=3.5cm, width=0.75\linewidth]{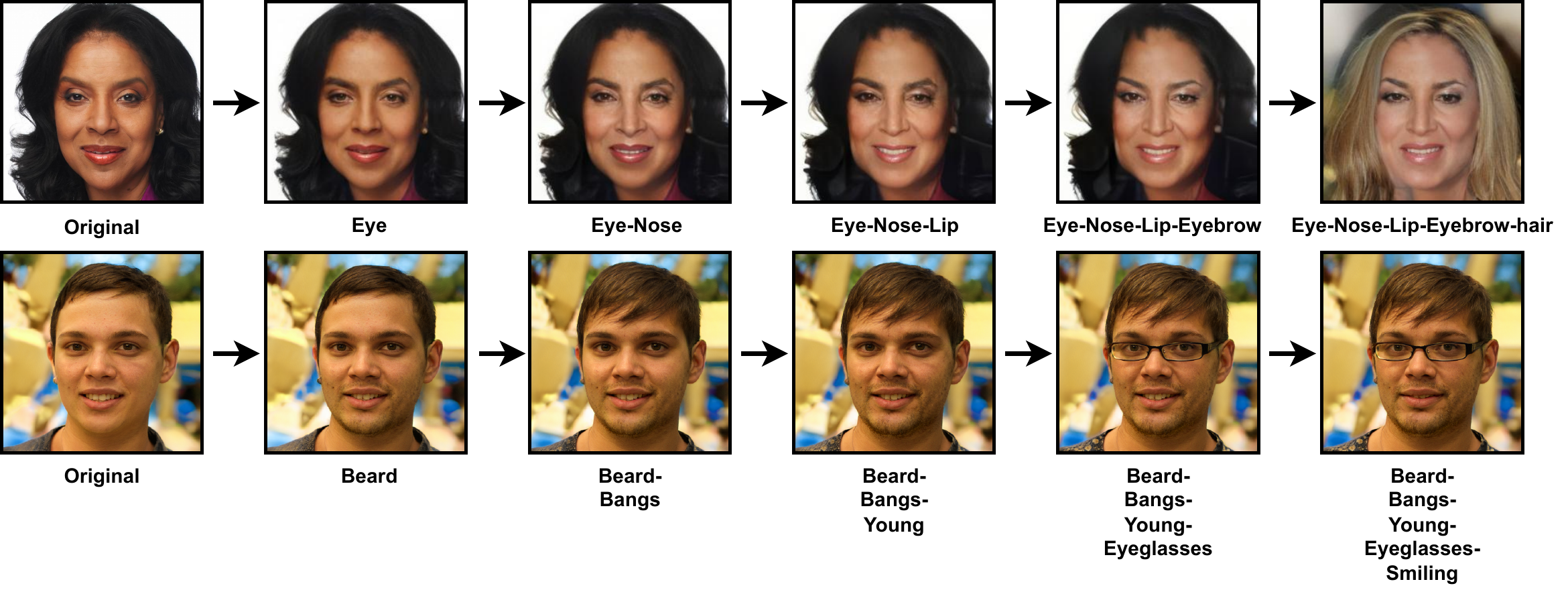}
	\end{center}
	\vspace*{-8mm}
	\caption{Illustration of sequential facial manipulation. Two types of facial manipulation approaches are considered, \textit{i.e.} facial components manipulation~\cite{kim2021exploiting} in the first row and facial attributes manipulation~\cite{jiang2021talk} in the second row.}
	\label{fig:DatasetIntro}
	\vspace{-20pt}
\end{figure}

To facilitate the study of detecting Seq-Deepfake, this paper contributes the first Seq-Deepfake dataset. Fig.~\ref{fig:DatasetIntro} shows some samples in Seq-Deepfake dataset. From Fig.~\ref{fig:DatasetIntro}, it can be seen that one face image can be sequentially manipulated with different number of steps (from minimum 1 step to maximum 5 steps), which leads to facial manipulation sequences with diverse lengths. It is extremely difficult to distinguish the original and manipulated face images, and even harder to figure out the exact manipulation sequences. To make our study more comprehensive, we consider two different facial manipulation techniques, facial components manipulation~\cite{kim2021exploiting} and facial attributes manipulation~\cite{jiang2021talk}, which are displayed in the first and second row, respectively in Fig.~\ref{fig:DatasetIntro}.

Most current facial manipulation applications are built based on Generative Adversarial Network (GAN). It is well known that the semantic latent space learned by GAN is difficult to be perfectly disentangled~\cite{shen2020interpreting,lee2020high}. We argue that this defect is likely to leave some spatial as well as sequential manipulation traces unveiling sequential facial manipulations. Based on this observation, to 
detect such two types of manipulations traces, we cast detecting Seq-Deepfake as a specific image-to-sequence (\textit{e.g.} image captioning) task and thus propose a concise yet effective Seq-DeepFake Transformer (\textbf{SeqFakeFormer}). Two key parts are devised in SeqFakeFormer: \textbf{Spatial Relation Extraction} and \textbf{Sequential Relation Modeling with Spatially Enhanced Cross-attention}. Given a manipulated image, to adaptively capture subtle spatial manipulation regions, SeqFakeFormer feeds the image into a deep convolutional neural network (CNN) to learn its feature maps. Then we extract the relation of spatial manipulations captured in feature maps using the self-attention modules of transformer encoder, obtaining features of spatial relation, i.e. spatial manipulation traces. After that, the decoder of SeqFakeFormer models the sequential relation of extracted features of spatial relation via cross-attention modules in an autoregressive mechanism, contributing to the detection of sequential manipulation traces, and thus detecting the facial manipulation sequences. To enable more effective cross-attention given limited annotations of facial manipulation sequences in Seq-DeepFake, SeqFakeFormer further integrates a Spatially Enhanced Cross-Attention (SECA) module in the decoder. This module enriches the spatial information of annotations of manipulation sequences by learning a spatial weight map. After fusing the spatial weight map with the cross-attention map, a spatially enhanced cross-attention can be achieved.

Main contributions of our paper can be summarized as follows:

\begin{itemize}
\item We introduce a new research problem named Detecting Sequential Deepfake Manipulation (\textbf{Seq-DeepFake}), with the objective of detecting sequences of facial manipulations, which expands the scope and poses a new challenge for deepfake detection.

\item We contribute the Sequential Deepfake Dataset with sequential manipulated face images using two different facial manipulation techniques. Corresponding annotations of manipulation sequences are provided.

\item We propose a powerful Seq-DeepFake Transformer (\textbf{SeqFakeFormer}). A comprehensive benchmark is built and rigorous evaluation protocols and metrics are designed for this novel research problem. Extensive quantitative and qualitative experiments demonstrate its superiority.
\end{itemize}

\section{Related Work}
\noindent \textbf{Deepfake detection.} Current deepfake detection methods can be roughly categorized into spatial-based and frequency-based deepfake detection. The majority of spatial-based deepfake detection methods focus on capturing visual cues from spatial domain. Face X-ray~\cite{li2020face} is proposed to detect the blending boundary left in the face forgery process as visual cues for real/fake detection. A multi-attentional deepfake detection network is proposed in~\cite{zhao2021multi} to integrate low-level textural features and high-level semantic features. Zhu \textit{et al.}~\cite{zhu2021face} introduce 3D decomposition into forgery detection and propose a two-stream network to fuse decomposed features for detection. Pair-wise self-consistency learning (PCL)~\cite{zhao2021learning} is introduced to detect inconsistency of source features within the manipulated images. Inconsistencies in semantically high-level mouth movements are captured in ~\cite{haliassos2021lips} by fine-tuning a temporal network pretrained on lipreading. On the other hand, some methods pay attention to the frequency domain for detecting spectrum artifacts. There exist distinct spectrum distributions and characteristics between real and fake images in the high-frequency part of Discrete Fourier Transform (DFT)~\cite{durall2019unmasking,dzanic2020fourier}. Qian \textit{et al.}~\cite{qian2020thinking} propose a F$^{3}$-Net to learn local frequency statistics based on Discrete Cosine Transform (DCT) to mine forgery. Liu \textit{et al.}~\cite{liu2021spatial} present a Spatial-Phase Shallow Learning method to fuse spatial image and phase spectrum for the up-sampling artifacts detection. A two-stream model is devised in~\cite{luo2021generalizing} to model the correlation between extracted high-frequency features and regular RGB features to learn generalizable features. A frequency-aware discriminative feature learning framework~\cite{li2021frequency} is introduced to integrate metric learning and adaptive frequency features learning for face forgery detection.

So far, several deepfake datasets have been released to public, such as FaceForensics++~\cite{rossler2019faceforensics++}, Celeb-DF~\cite{li2020celeb}, Deepfake Detection Challenge (DFDC)~\cite{dolhansky2019deepfake}, and DeeperForensics-1.0 (DF1.0)~\cite{jiang2020deeperforensics}. However, only binary labels are provided in most of existing deepfake datasets, and thus most of the above works are trained to carry out binary classification, which results in performance saturation and poor generalization.

\noindent \textbf{Facial editing.}
Several methods have been proposed for editing facial components (\textit{i.e.} eye, nose, month). Lee \textit{et al.}~\cite{lee2020maskgan} present a geometry-oriented face manipulation network MaskGAN for diverse and interactive face manipulation guided by semantic masks annotations. A semantic region-adaptive normalization (SEAN)~\cite{zhu2020sean} is proposed to facilitate manipulating face images by encoding images into the per-region style codes conditioned on segmentation masks. StyleMapGAN~\cite{kim2021exploiting} introduces explicit spatial dimensions to the latent space and manipulates facial components by blending the latent spaces between reference and original faces. Moreover, some works target editing specific facial attributes such as age progression~\cite{yang2018learning}, and smile generation~\cite{wang2018every}. Some recent works discover semantically meaningful directions in the latent space of a pretrained GAN so as to carry out facial attributes editing by moving the latent code along these directions~\cite{shen2020interpreting,shen2020interfacegan,zhuang2021enjoy,voynov2020unsupervised,shen2021closed}. InterFaceGAN~\cite{shen2020interpreting,shen2020interfacegan} tries to disentangle attribute representations in the latent space of GANs by searching a hyperplane, of which a normal vector is used as the editing direction. Fine-grained facial attributes editing is achieved by~\cite{jiang2021talk} through searching a curving trajectory with respect to attribute landscapes in the latent space of GANs.

\section{Sequential Deepfake Dataset}

\begin{figure}[t] 
	\begin{center}
		\includegraphics[height=5cm, width=0.93\linewidth]{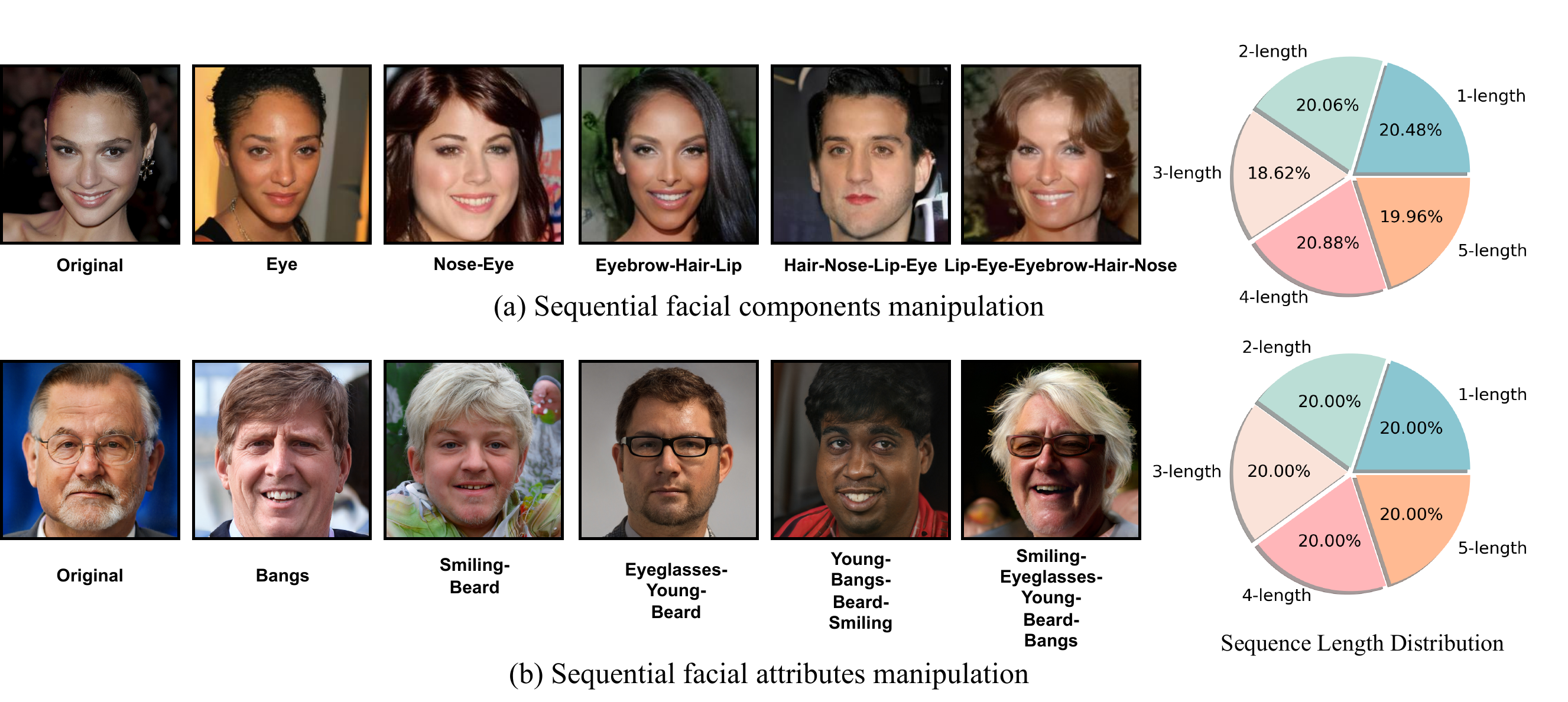}
	\end{center}
	\vspace*{-7mm}
	\caption{Illustration of Seq-Deepfake dataset. Samples of Seq-Deepfake are provided with annotations of manipulation sequences. We also show sequence length distribution.}
	\label{fig:DatasetExp}
	\vspace{-16pt}
\end{figure}

To support the novel research problem, we generate a large-scale Sequential Deepfake (Seq-Deepfake) dataset consisting of sequential manipulated face images based on two representative facial manipulation techniques, facial components manipulation~\cite{kim2021exploiting} and facial attributes manipulation~\cite{jiang2021talk}. Unlike most of existing deepfake datasets~\cite{rossler2019faceforensics++,he2021forgerynet} only providing binary labels, the proposed dataset contains annotations of manipulation sequences with diverse sequence lengths. Details of generation pipelines based on the two facial manipulation techniques are as follows.

\noindent \textbf{Sequential facial components manipulation.}
We adopt the StyleMapGAN proposed in~\cite{kim2021exploiting} for facial components manipulation. Facial components manipulation is carried out based on original images from CelebA-HQ~\cite{liu2015deep,karras2017progressive} and corresponding facial component masks from CelebAMask-HQ~\cite{lee2020maskgan} dataset. Facial components manipulation aims to transplant some facial components of a reference image to an original image with respect to a mask that indicates the components to be manipulated. Specifically, we project the original image and the reference image through the encoder of StyleMapGAN to obtain stylemaps, which are intermediate latent spaces with spatial dimensions. Then, the facial components manipulation is carried out by blending the stylemaps extracted from reference and original faces based on facial component masks. Due to the inevitable appearance of degraded images in the generation process, we adopt the Generated Image Quality Assessment (GIQA) algorithm~\cite{gu2020giqa} to quantitatively evaluate the quality of each generated image and then filter out some low-quality ones based on the pre-defined threshold. Fig.~\ref{fig:DatasetExp} (a) shows some samples with corresponding annotations of sequential facial components manipulation. Through this data generation pipeline, we can finally generate 35,166 manipulated face images annotated with 28 types of manipulation sequences in different lengths (including original). As illustrated in Fig.~\ref{fig:DatasetExp} (a), the proportions of 1-5 different lengths of manipulation sequences are: 20.48\%, 20.06\%, 18.62\%, 20.88\%, 19.96\%.

\noindent \textbf{Sequential facial attributes manipulation.} Unlike facial components manipulation methods that swap certain local parts from a reference image to an original image, facial attributes manipulation approaches directly change specific attributes on the original face image without any reference images. To take this manipulation type into consideration, we utilize the fine-grained facial editing method proposed by~\cite{jiang2021talk}. This method aims to learn a location-specific semantic field for each editing type on the training set, then edit this attribute of interest on the given face image to a user-defined degree by stepping forward or backward on the learned curve in latent space. Based on this idea, we further generate face images with sequential facial attributes manipulation by performing the editing process in a sequential manner. Specifically, we first sample latent codes from the StyleGAN trained on FFHQ dataset~\cite{karras2019style} to generate original images. Then according to pre-defined attribute sequences, we progressively manipulate each attribute on the original face to another randomly chosen degree using the above method. After generating the final manipulation results, we also perform GIQA algorithm to filter out low-quality samples. Using this pipeline, we generate 49,920 face images with 26 manipulation sequence types, with the length of each sequence ranging from 1 to 5. Since this generation pipeline is more controllable than facial components manipulation, we construct a more balanced dataset, as shown in Fig.~\ref{fig:DatasetExp} (b).

\section{Our Approach}

\begin{figure}[t] 
	\begin{center}
		\includegraphics[height=2.3cm, width=0.9\linewidth]{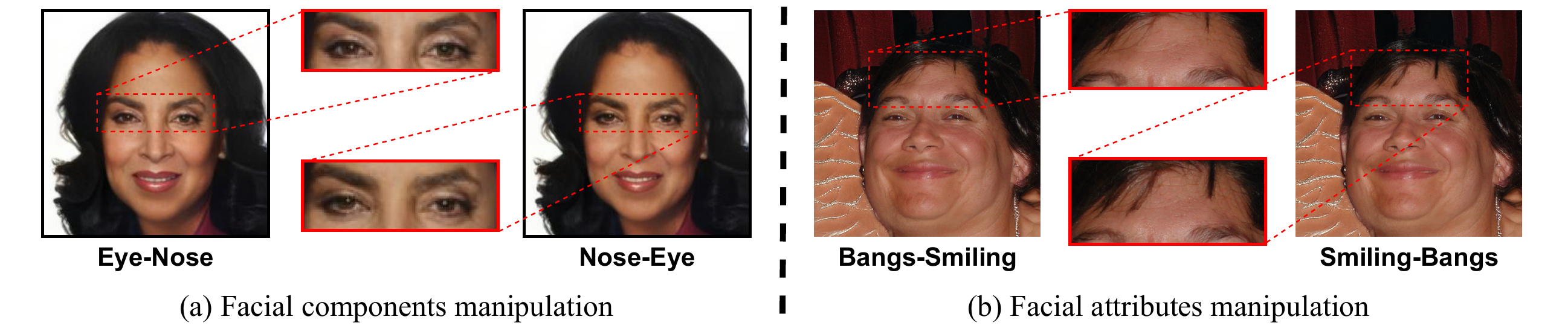}
	\end{center}
	\vspace*{-7mm}
	\caption{Effect of different sequential order for facial manipulation. Switching the sequential order of manipulations between (a) eye and nose and (b) bangs and smiling results in different facial manipulations.}
	\label{fig:order}
	\vspace{-16pt}
\end{figure}

\subsection{Motivation}

Most current facial manipulation applications are constructed using algorithms of Generative Adversarial Network (GAN). However, it is a well known fact that due to imperfect semantic disentanglement in the latent space of GAN~\cite{shen2020interpreting,lee2020high}, manipulating one facial component or attribute is likely to affect the others. As shown in the first row of Fig.~\ref{fig:DatasetIntro}, manipulating the nose in the step of `Eye-Nose' simultaneously results in some little modification on the eye and mouth components compared to the former step `Eye', which alters the overall \textbf{spatial relation} among facial components. We can thus discover some \textbf{spatial manipulation traces} from the spatial relation. Furthermore, as illustrated in Fig.~\ref{fig:order}, switching the sequential order of manipulations (\textit{e.g.,} manipulation order between eye and nose in (a) and bangs and smiling in (b) in Fig.~\ref{fig:order}) causes different facial manipulation results (\textit{e.g.,} distinct gazes in (a) and distinct amount of bangs in (b) in Fig.~\ref{fig:order}), which indicates that when changing the sequential order of manipulations, the above overall spatial relation of facial components altered by manipulations will also be changed. This means there exists sequential information from spatial relation that reflects the sequential order of manipulations, which corresponds to the facial manipulation sequence. That is, we can extract the spatial relation among facial components to unveil the \textbf{spatial manipulation traces} and model their \textbf{sequential relation} to detect the facial manipulation sequence. We thus regard the sequential relation as \textbf{sequential manipulation traces}.



\subsection{Overview}
Based on the above observation, we cast detecting Seq-Deepfake manipulation as a specific image-to-sequence task, where inputs are manipulated/original images and outputs are facial manipulation sequences. Three challenges will be encountered when addressing the task. 1) From Fig.~\ref{fig:DatasetIntro} and \ref{fig:DatasetExp}, it can be seen that distinguishing original faces and sequential manipulated faces is extremely hard. Besides, with respect to different people, differences in face contour cause diverse manipulation regions for the same type of facial components/attributes manipulation. Thus, given indistinguishable and diverse facial manipulations, how to adaptively capture subtle manipulation regions and model their spatial relation accurately is quite challenging. 2) Based on the spatial relation of manipulated components/attributes, how to precisely model their sequential relation so as to detect the sequential facial manipulation is another challenge. 3) Compared to normal image-to-sequence task (\textit{e.g.} image captions), the annotations of manipulation sequences are much shorter and thus less informative in our task. Therefore, how to effectively learn the sequential information of facial manipulations given limited annotations of manipulation sequences should also be considered.

\begin{figure}[t] 
	\begin{center}
		\includegraphics[height=4.2cm, width=1\linewidth]{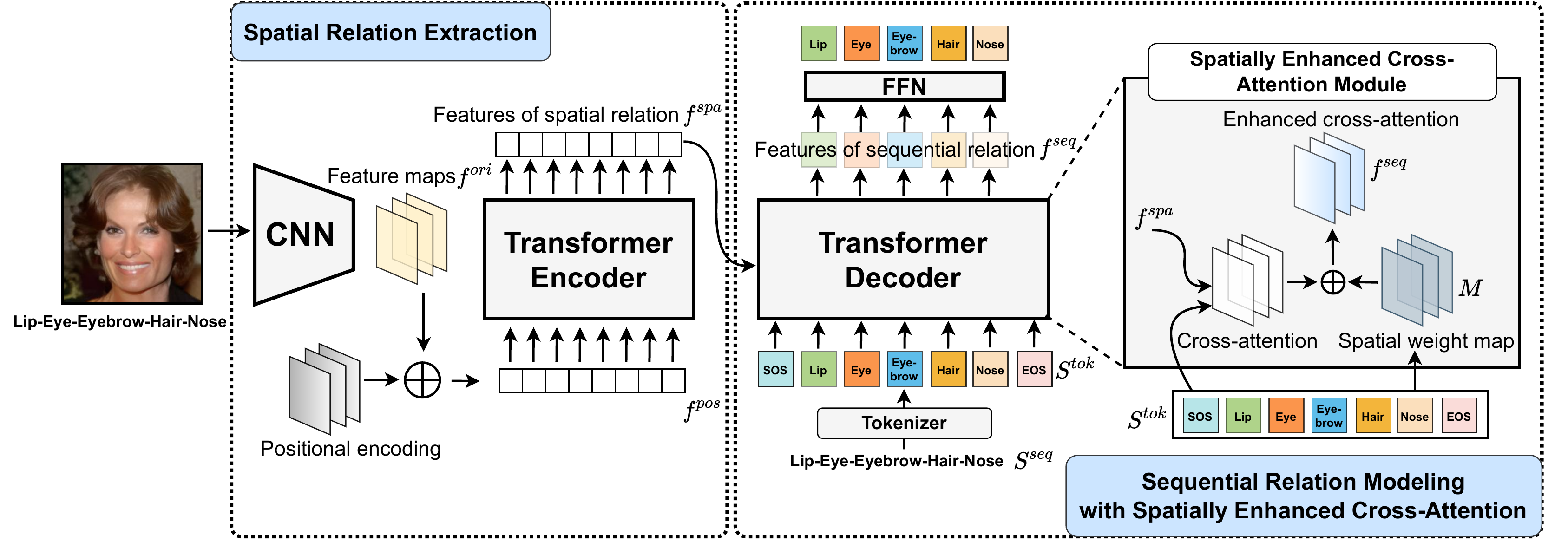}
	\end{center}
	\vspace*{-6mm}
	\caption{Overview of proposed Seq-DeepFake Transformer (\textbf{SeqFakeFormer}). We first feed the face image into a CNN to learn features of spatial manipulation regions, and extract their spatial relation via self-attention modules in the encoder. Then sequential relation based on features of spatial relation is modeled to detect the sequential facial manipulation. A spatial enhanced cross-attention module is integrated into the decoder, contributing to a more effective cross-attention.}
	\label{fig:transformer}
	\vspace{-16pt}
\end{figure}

To cope with the above three challenges, as shown in Fig.~\ref{fig:transformer}, we propose a  Seq-DeepFake  Transformer (\textbf{SeqFakeFormer}), which is composed of two key parts: \textbf{Spatial Relation Extraction}, \textbf{Sequential Relation Modeling with Spatially Enhanced Cross-attention}. To capture spatial manipulation traces, features of subtle manipulation regions are first adaptively captured by a CNN and their spatial relation are extracted via self-attention modules in the transformer encoder. After that, we capture sequential manipulation traces by modeling sequential relation based on features of spatial relation through cross-attention modules deployed in the decoder with an auto-regressive mechanism. To achieve more effective cross-attention given limited annotations of manipulation sequences, a spatially enhanced cross-attention module is devised to generate different spatial weight maps for corresponding manipulations to carry out cross-attention. In the following subsections, we describe all components in detail.

\subsection{Spatial Relation Extraction}

To adaptively capture subtle and various facial manipulation regions, we exploit a CNN to learn feature maps of the input image. Given an input image $x$ $\in$ R$^{3 \times H' \times W'}$, we first feed it into a CNN~\cite{he2016deep} to extract its visual feature maps $f^{ori}$ = CNN($x$), $f^{ori} \in$ R$^{C \times  H \times W}$, where $H', W'$, and $H, W$ are the height and width of the input image and its corresponding feature maps, respectively. $C$ is the number of channels of feature maps.

Since the transformer architecture is permutation-invariant, we supplement original visual features maps $f^{ori}$ with fixed positional encodings~\cite{parmar2018image,bello2019attention}, resulting in feature maps denoted as $f^{pos}$. Since transformer encoder accepts a sequence as input, we reshape the spatial dimensions of $f^{pos}$ to one dimension, generating reshaped features $f^{pos} \in$ R$^{C \times HW}$. After feeding into the transformer encoder, $f^{pos}$ conducts self-attention by generating the key, query, and value features $K, Q, V$ so as to extract the relations among all spatial positions. Through this self-attention operation on CNN features, spatial relation of manipulation regions are exploited and thus spatial manipulation traces can be extracted. To further facilitate spatial relation extraction, this paper adopts multi-head self-attention which splits features $f^{pos}$ into multiple groups along the channel dimension. The multi-head normalized attention based on dot-product is as follows:
\begin{equation}
\begin{split}
& f^{spa}_i = \text{Softmax}(K_i^T Q_i / \sqrt{d})V_i, f^{spa} = \text{Concat}(f^{spa}_1, ... , f^{spa}_D)
\end{split}
\end{equation}

\noindent where $K_i , Q_i , V_i$ denote the $i$-th group of the key, query, and value features, $d$ is dimension of queries and keys, and total $D$ groups are generated. We then concatenate all the groups to form the features of spatial relation $f^{spa}$ as the output of encoder. 

\subsection{Sequential Relation Modeling with Spatially Enhanced Cross-Attention}

Given features of spatial relation $f^{spa}$ extracted from the encoder, we propose to model the sequential relation among them to detect the facial manipulation sequences. To this end, we carry out cross-attention between features of spatial relation $f^{spa}$ and corresponding annotations of manipulation sequences in an auto-regressive manner. To achieve this, we send original annotations of manipulation sequences $S^{ori} \in R^{C \times N}$ (\textit{e.g.,} $N$=5 in Fig.~\ref{fig:transformer} before a Tokenizer) into a Tokenizer, where we transform each manipulation in the sequence into one token and insert Start of Sentence (SOS) and End of Sentence (EOS) tokens into the beginning and end of sequence. After that, we obtain tokenized manipulation sequences $S^{tok} \in R^{C \times (N+2)}$ to be cross-attended with features of spatial relation $f^{spa}$. With the auto-regressive mechanism, the decoding process of facial manipulation sequence in the transformer decoder (aided by cross-attention) is triggered by SOS token and will be automatically stopped once the EOS token is predicted. In this way, we can predict facial manipulation sequences with adaptive lengths.

Normally, cross-attention between tokenized sequences $S^{tok}$ and features of spatial relation $f^{spa}$ should be performed directly. However, as mentioned above, compared to the normal image-to-sequence task, annotations of manipulation sequences are much shorter and thus less informative ($S^{tok}$ only has $(N+2)$-length and maximum of $N$ is 5). To effectively cross-attend features of spatial relation with limited annotations of manipulation sequences, inspired by~\cite{gao2021fast}, we propose a sequential relation modeling with spatially enhanced cross-attention. We argue that each manipulation in $S^{tok}$ corresponds to one specific facial component/attribute which has a strong prior of spatial regions, thus we can enrich the information of manipulation sequences guided by this prior. To this end, we generate the spatial weight map for each manipulation by dynamically predicting the spatial center and scale of each manipulation component/attribute in annotations of manipulation sequences as follows:

\begin{equation}
\begin{split}
& t_h, t_w =  \text{sigmoid}(\text{MLP}(S^{tok})), r_h, r_w = \text{FC}(S^{tok}) \\
\end{split}
\label{equ:centerscale}
\end{equation}

\noindent where $t_h, t_w$ and $r_h, r_w$ are estimated 2-dimensional coordinates corresponding to spatial centers and scales of specific manipulations in the sequences, respectively. Then the Gaussian-shape spatial weight map can be generated as:

\begin{equation}
\begin{split}
& M(h, w) =  \text{exp}\left(-\frac{(h-t_h)^2}{\lambda r_h^2}-\frac{(w-t_w)^2}{\lambda r_w^2} \right)
\end{split}
\label{equ:spatialmap}
\end{equation}

\noindent where $(h, w) \in [0, H] \times [0, W]$ are 2-dimensional coordinates of the spatial weight map $M$, and $\lambda$ is a hyper-parameter to modulate the bandwidth of the Gaussian-shape distribution. From Eq.~\ref{equ:spatialmap}, it can be seen that spatial weight map $M$ can assign higher importance to spatial regions closer to the centers and lower weights to locations farther from the centers. Moreover, as analyzed before, since diverse manipulation regions are presented for different people, the above dynamically learned scales can further tune the height/width ratios of spatial weight map based on each manipulation, contributing to a more adaptive spatial weight map. Based on this idea, we can enhance the cross-attention between features of spatial relation and annotations of manipulation sequences with generated spatial weight map $M$ as follows:

\begin{equation}
\begin{split}
& S = \text{FC}(S^{tok}), K, V = \text{FC}(f^{spa}), \\
& f^{seq}_i = \text{Softmax}(K_i^T Q_i \sqrt{d} + logM)V_i, \\
& f^{seq} = \text{Concat}(f^{seq}_1, ..., f^{seq}_D)
\end{split}
\label{equ:decoder}
\end{equation}

\noindent where FC denotes a single fully-connected layer, and $f^{seq}_i$ denotes features of sequential relation. The cross-attention of the $i$-th head is further element-wise added with logarithm of spatial weight map $M$, which contributes to spatially enhanced cross-attention. Furthermore, to model the sequential relation of facial manipulation, the auto-regressive mechanism is integrated into the above cross-attention process. This is implemented by masking out (setting to $-\infty$) all values in the input of the Softmax function in Eq.~\ref{equ:decoder} which correspond to illegal connections. Through concatenation of features of sequential relation from all cross-attention heads, we can obtain the final features of sequential relation $f^{seq}$ as the output of decoder.

The features of sequential relation are then fed into a Fast Forward Network (FFN) and transformed to a class score for each manipulation. Finally, we jointly train the CNN, transformer encoder and decoder by minimizing the cross-entropy loss between each class score and corresponding annotation of manipulation in the sequence.

\section{Experiments}
\subsection{Experimental Setup}

\noindent \textbf{Implementation Details.} 
We choose two different CNNs, ResNet-34~\cite{he2016deep} and ResNet-50~\cite{he2016deep} pre-trained on ImageNet~\cite{deng2009imagenet} dataset in our paper. To be comparable in the number of parameters, we adopt a transformer model with 2  encoder and 2 decoder layers with 4 attention heads. For the training schedule, we employ 20 epochs warm-up strategy and train for 170 epochs with a learning rate drop to 10\% in every 50 epochs. The initial learning rates are set as 1$e-3$ for transformer part and 1$e-4$ for CNN part. We set $\lambda=4$.

\noindent \textbf{Baseline Methods.} 
The most straightforward solution for detecting Seq-Deepfake manipulation is to regard it as a multi-label classification problem~\cite{wang2021can}. It treats all manipulations in the sequences as independent classes and classifies the manipulated images into multiple manipulation classes. Specifically, we design a simple multi-label classification network (denoted as \textbf{Multi-Cls}) as one of the baselines. We use ResNet-34~\cite{he2016deep} and ResNet-50~\cite{he2016deep} pre-trained on ImageNet~\cite{deng2009imagenet} dataset as backbones for the multi-label classification network, which is concatenated with $N$ single linear-layer branches as $N$ classification heads ($N=5$ as maximum manipulation steps are 5 in Seq-Deepfake dataset). Moreover, we study a more complex transformer structure modified for our problem. \textbf{DETR}~\cite{carion2020end} is a popular transformer devised for end-to-end object detection. This model detects input images' bounding boxes and corresponding object classes conditioned on object queries. We revise this model by replacing the object queries with annotations of manipulation sequences and only preserve the output of object classes. 
Furthermore, to examine the performance of existing deepfake detection methods for our research problem, we adapt three state-of-the-art deepfake detection methods, a Dilated Residual Network variant (\textbf{DRN})~\cite{wang2019detecting}, a two-stream network modeling the correlation between high-frequency features and regular RGB features (\textbf{Two-Stream})~\cite{luo2021generalizing}, and a multi-attentional deepfake detection (\textbf{MA})~\cite{zhao2021multi}, into multi-label classification setting. To be specific, we replace their binary label classifier with multiple classification heads to classify sequential manipulations. Please note since all of the above baselines are only able to predict the facial manipulation with fixed length ($N=5$), `no manipulation' class will be padded into the annotation sequence if its length is shorter than $N$ so that we can keep the same length between predictions and annotation sequences for training. 

\noindent \textbf{Evaluation Metrics.} We propose two evaluation metrics for this new task.
\begin{itemize}
\item \textbf{Fixed Accuracy (Fixed-Acc):}  Given prediction with fixed $N$-length ($N=5$) by above baselines, as in the training process, the first type of evaluation pads `no manipulation' class into the annotated manipulation sequences and compares each manipulation class in the predicted sequences with its corresponding annotation to calculate the evaluation accuracy.

\item \textbf{Adaptive Accuracy (Adaptive-Acc):} Moreover, since the proposed method exploits sequential information to detect facial manipulation sequences based on the auto-regressive mechanism, predictions will be automatically stopped once predicting the EOS token. Thus, the proposed method can detect facial manipulation sequences with adaptive lengths. To conduct the evaluation in this scenario, the second type of evaluation is devised, which compares predicted manipulations and corresponding annotations within the maximum steps of manipulations ($N \leq 5$) between them. This makes the evaluation focus more on accuracy of manipulations.
\end{itemize}
More details of two evaluation metrics can be found in \textbf{Supplementary Material}. 

\subsection{Benchmark for Seq-Deepfake}

\begin{table}[t]
\scriptsize
\renewcommand{\arraystretch}{1}
\centering
\caption{Accuracy of detecting Seq-Deepfake based on sequential facial components manipulation}
\begin{tabular}{c|cc|cc}
\Xhline{2.5\arrayrulewidth}
\multirow{2}{*}{Methods} & \multicolumn{2}{c|}{ResNet-34}                       & \multicolumn{2}{c}{ResNet-50}                        \\ \cline{2-5} 
                         & \multicolumn{1}{c|}{Fixed-Acc}      & Adaptive-Acc   & \multicolumn{1}{c|}{Fixed-Acc}      & Adaptive-Acc   \\ \hline
Multi-Cls                & \multicolumn{1}{c|}{69.66}          & 50.54          & \multicolumn{1}{c|}{69.65}          & 50.57          \\
DETR~\cite{carion2020end}                     & \multicolumn{1}{c|}{69.87}          & 50.63          & \multicolumn{1}{c|}{69.75}          & 49.84          \\
Ours                     & \multicolumn{1}{c|}{\textbf{72.13}} & \textbf{54.80} & \multicolumn{1}{c|}{\textbf{72.65}} & \textbf{55.30} \\ \Xhline{2.5\arrayrulewidth}
\end{tabular}
\label{tbl:facecomp}
\vspace{-16pt}
\end{table}

\begin{table}[t]
\scriptsize
\renewcommand{\arraystretch}{1}
\centering
\caption{Accuracy of detecting Seq-Deepfake based on sequential facial attributes manipulation}
\begin{tabular}{c|cc|cc}
\Xhline{2.5\arrayrulewidth}
\multirow{2}{*}{Methods} & \multicolumn{2}{c|}{ResNet-34}                       & \multicolumn{2}{c}{ResNet-50}                        \\ \cline{2-5} 
                         & \multicolumn{1}{c|}{Fixed-Acc}      & Adaptive-Acc   & \multicolumn{1}{c|}{Fixed-Acc}      & Adaptive-Acc   \\ \hline
Multi-Cls                & \multicolumn{1}{c|}{66.99}          & 46.68          & \multicolumn{1}{c|}{66.66}          & 46.00          \\
DETR~\cite{carion2020end}                     & \multicolumn{1}{c|}{67.93}          & 48.15          & \multicolumn{1}{c|}{67.62}          & 47.99          \\
Ours                     & \multicolumn{1}{c|}{\textbf{67.99}} & \textbf{48.32} & \multicolumn{1}{c|}{\textbf{68.86}} & \textbf{49.63} \\ \Xhline{2.5\arrayrulewidth}
\end{tabular}
\label{tbl:faceattr}
\vspace{-16pt}
\end{table}

\begin{table}[t]
\scriptsize
\renewcommand{\arraystretch}{1}
\centering
\caption{Accuracy of detecting Seq-Deepfake compared to deepfake detection methods}
\begin{tabular}{c|cc|cc}
\Xhline{2.5\arrayrulewidth}
\multirow{2}{*}{Methods} & \multicolumn{2}{c|}{Face Components Manipulation}    & \multicolumn{2}{c}{Face Attributes Manipulation}     \\ \cline{2-5} 
                         & \multicolumn{1}{c|}{Fixed-Acc}      & Adaptive-Acc   & \multicolumn{1}{c|}{Fixed-Acc}      & Adaptive-Acc   \\ \hline
DRN~\cite{wang2019detecting}               & \multicolumn{1}{c|}{66.06}          & 45.79          & \multicolumn{1}{c|}{64.42}          & 43.20          \\
MA~\cite{zhao2021multi}                     & \multicolumn{1}{c|}{71.31}          & 52.94          & \multicolumn{1}{c|}{67.58}          & 47.48      \\
Two-Stream~\cite{luo2021generalizing}               & \multicolumn{1}{c|}{71.92}          & 53.89          & \multicolumn{1}{c|}{66.77}          & 46.38          \\
Ours                     & \multicolumn{1}{c|}{\textbf{72.65}} & \textbf{55.30} & \multicolumn{1}{c|}{\textbf{68.86}} & \textbf{49.63} \\ \Xhline{2.5\arrayrulewidth}
\end{tabular}
\label{tbl:deepfakebaseline}
\vspace{-16pt}
\end{table}

We tabulate the first benchmark for detecting sequential facial manipulation based on facial components manipulation and facial attributes manipulation in Tables~\ref{tbl:facecomp} to \ref{tbl:deepfakebaseline}. We note that, both baselines and the proposed method obtains much higher performance under evaluation metric Fixed-Acc than Adaptive-Acc. This validates that detecting sequential facial manipulation with adaptive lengths is much harder than its simplified version with fixed length. It can be observed from Tables~\ref{tbl:facecomp} and \ref{tbl:faceattr}, that the proposed SeqFakeFormer obtains the best performance of detecting facial manipulation sequences compared to all considered baselines in both facial components manipulation and facial attributes manipulation. In addition, SeqFakeFormer also performs better than other baselines with both CNNs (ResNet-34 and ResNet-50), indicating the compatibility of the proposed method with different feature extractors. Specifically, the proposed method has achieved about 3-4\% improvement in facial components sequential manipulation and 1-2\% improvement in facial attributes sequential manipulation under two evaluation metrics. In particular, there exists a larger performance gap between SeqFakeFormer and other baselines under evaluation metric Adaptive-Acc than Fixed-Acc, which demonstrates that the effectiveness of the proposed method is more significant in the harder case. Moreover, we tabulate the comparison between three SOTA deepfake detection methods and our method in Table~\ref{tbl:deepfakebaseline}. SeqFakeFormer also outperforms all SOTA deepfake detection methods in both manipulation types. Since all the baselines treat detecting Seq-Deepfake as a multi-label classification problem, only spatial information of manipulated images are extracted. In contrast, SeqFakeFormer is capable of exploiting both spatial and sequential manipulation traces and thus more useful sequential information can be modeled, which is the key to enhance the performance of Seq-Deepfake Detection.

\subsection{Ablation study}

\begin{table}[t]
\scriptsize
\renewcommand{\arraystretch}{1}
\centering
\caption{Ablation study of detecting Seq-Deepfake based on sequential facial components manipulation}
\begin{tabular}{cc|cc|cc}
\Xhline{2.5\arrayrulewidth}
\multicolumn{2}{c|}{Components}                      & \multicolumn{2}{c|}{ResNet-34}                       & \multicolumn{2}{c}{ResNet-50}                        \\ \hline
\multicolumn{1}{c|}{Auto-regressive} & SECA          & \multicolumn{1}{c|}{Fixed-Acc}      & Adaptive-Acc   & \multicolumn{1}{c|}{Fixed-Acc}      & Adaptive-Acc   \\ \hline
\multicolumn{1}{c|}{\XSolidBrush}     & \XSolidBrush   & \multicolumn{1}{c|}{70.64}          & 52.19          & \multicolumn{1}{c|}{71.22}          & 53.43          \\
\multicolumn{1}{c|}{\XSolidBrush}     & \CheckmarkBold & \multicolumn{1}{c|}{70.77}          & 51.71          & \multicolumn{1}{c|}{70.99}          & 52.66          \\
\multicolumn{1}{c|}{\CheckmarkBold}   & \XSolidBrush   & \multicolumn{1}{c|}{71.88}          & 53.84          & \multicolumn{1}{c|}{72.18}          & 54.64          \\
\multicolumn{1}{c|}{\CheckmarkBold}   & \CheckmarkBold & \multicolumn{1}{c|}{\textbf{72.13}} & \textbf{54.80} & \multicolumn{1}{c|}{\textbf{72.65}} & \textbf{55.30} \\ \Xhline{2.5\arrayrulewidth}
\end{tabular}
\label{tbl:Ablfacecomp}
\vspace{-16pt}
\end{table}

\begin{table}[t]
\scriptsize
\renewcommand{\arraystretch}{1}
\centering
\caption{Ablation study of detecting Seq-Deepfake based on sequential facial attributes manipulation}
\begin{tabular}{cc|cc|cc}
\Xhline{2.5\arrayrulewidth}
\multicolumn{2}{c|}{Components}                      & \multicolumn{2}{c|}{ResNet-34}                       & \multicolumn{2}{c}{ResNet-50}                        \\ \hline
\multicolumn{1}{c|}{Auto-regressive} & SECA          & \multicolumn{1}{c|}{Fixed-Acc}      & Adaptive-Acc   & \multicolumn{1}{c|}{Fixed-Acc}      & Adaptive-Acc   \\ \hline
\multicolumn{1}{c|}{\XSolidBrush}     & \XSolidBrush   & \multicolumn{1}{c|}{66.98}          & 45.87          & \multicolumn{1}{c|}{68.14}          & 48.49          \\
\multicolumn{1}{c|}{\XSolidBrush}     & \CheckmarkBold & \multicolumn{1}{c|}{67.36}          & 47.22          & \multicolumn{1}{c|}{68.77}          & 49.54          \\
\multicolumn{1}{c|}{\CheckmarkBold}   & \XSolidBrush   & \multicolumn{1}{c|}{66.70}          & 46.56          & \multicolumn{1}{c|}{68.17}          & 48.81          \\
\multicolumn{1}{c|}{\CheckmarkBold}   & \CheckmarkBold & \multicolumn{1}{c|}{\textbf{67.99}} & \textbf{48.32} & \multicolumn{1}{c|}{\textbf{68.86}} & \textbf{49.63} \\ \Xhline{2.5\arrayrulewidth}
\end{tabular}
\label{tbl:Ablfaceattr}
\end{table}

In this sub-section we investigate the impact of two key components in SeqFakeFormer, auto-regressive mechanism and Spatially Enhanced Cross-Attention module (SECA), to the overall performance. The considered components and the corresponding results obtained for each case are tabulated Tables~\ref{tbl:Ablfacecomp} and \ref{tbl:Ablfaceattr}. As evident from Tables~\ref{tbl:Ablfacecomp} and \ref{tbl:Ablfaceattr}, removing either auto-regressive mechanism or SECA will degrade the overall performance. This validates that auto-regressive mechanism facilitates the sequential relation modeling and SECA benefits the cross-attention. These components complement each other to produce better performance for detecting Seq-Deepfake.

\subsection{Face Recovery}

\begin{figure}[t]
    \centering
    \begin{minipage}[t]{0.75\textwidth}
    \centering
    \includegraphics[height=3.1cm, width=1\linewidth]{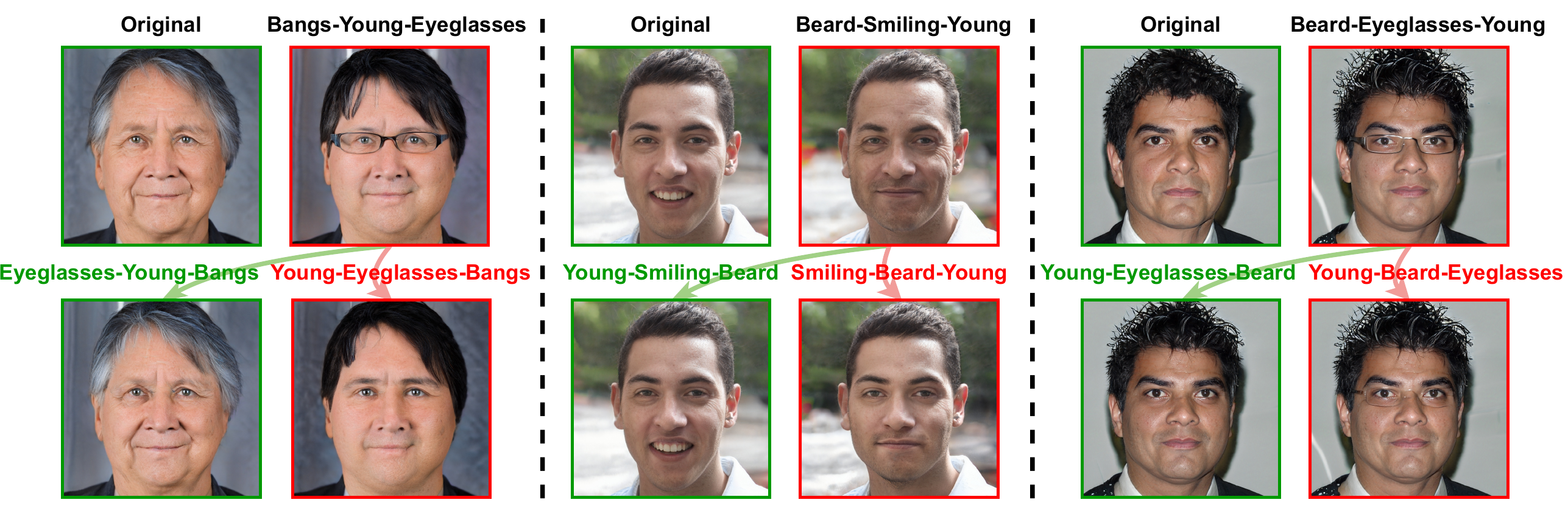}
    \caption{Face recovery based on correct and wrong facial manipulation sequences.}
    \label{fig:facerecover}
    \end{minipage}
    \begin{minipage}[t]{0.2\textwidth}
    \centering
    \includegraphics[height=3cm, width=1\linewidth]{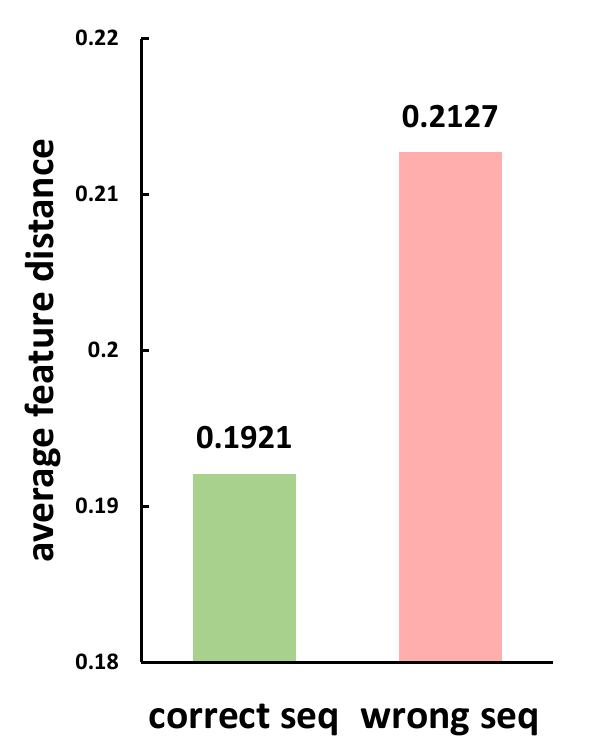}
    \caption{Identity preservation.}
    \label{fig:IDSocre}
    \end{minipage}
    \vspace{-16pt}
\end{figure}

After detecting facial manipulation sequences, we are able to perform more challenging tasks, like recovering the original face from the manipulated face image. Specifically, we formulate the Face Recovery task as: given a sequentially manipulated face image, reverse the manipulation process to get an image as close as possible to the original image. For example, in the facial attributes manipulation case, given an image generated by sequential manipulations on different attributes on the original face, we want to recover the original image. In fact, this task can be seen as an inverse sequential facial attribute manipulation problem, which can be effectively solved by the data generation pipeline described in Section 3 in an inverse manner. Specifically, as can be observed in Fig.~\ref{fig:facerecover}, once we detect the correct facial manipulation sequence, \textit{i.e.} correct manipulations ordered with correct manipulation steps, we can recover original face by performing face attribute manipulation based on the inverse order of detected facial manipulation sequence (process with green arrow). Comparatively, recovering the face image with wrongly ordered manipulation sequences may encounter different problems, such as incomplete recovery of age, smile, glasses, etc. (process with red arrow). Fig.~\ref{fig:IDSocre} evaluates the results using identity preservation metrics as in~\cite{jiang2021talk}, where smaller feature distance means identity is better preserved. The average feature distance between randomly selected 100 original faces and recovered faces using correct manipulation sequences is clearly smaller than that of the wrongly ordered sequence, indicating that the identity can be better recovered with correct manipulation sequence. Based on the above analysis and experiments, we prove that the detection of facial manipulation sequences is highly useful for face recovery, and we hope it can be applied to more meaningful tasks in the future.

\section{Conclusion}
This paper studies a novel research problem -- Detecting Sequential DeepFake Manipulation, aiming to detect a sequential vector of multi-step facial manipulation operations. We also introduce the first Seq-DeepFake dataset to provide sequentially manipulated face images. Supported by this new dataset, we cast detecting Seq-DeepFake manipulation as a specific image-to-sequence task and propose a Seq-DeepFake Transformer (SeqFakeFormer). Two modules, Spatial Relation Extraction and Sequential Relation Modeling with Spatially Enhanced Cross-Attention, are integrated into SeqFakeFormer, complementing each other. Extensive experimental results demonstrate the superiority of SeqFakeFormer and valuable observations pave the way for future research in broader deepfake detection.

%
%
\bibliographystyle{splncs04}
\bibliography{egbib}

\clearpage

\section*{Supplementary Material}
\appendix
	
\section{Training Details and Hyper-parameter Setting
}
Implementation is in PyTorch. For the training schedule, we employ a 20-epochs warm-up strategy. The initial learning rate is set as 1$e-3$ for transformer part and 1$e-4$ for CNN part, with a decay factor of 10 at 70 and 120 epochs, totally 170 epochs. We use the SGD momentum optimizer with weight decay set as $1e-4$. We use a mini-batch size of 32 per GPU and 4 GPUs in total. Model selection for evaluation is conducted by considering the trained model that has produced the best accuracy on the validation set.


\section{Evaluation Metrics}

\begin{figure}[] 
	\begin{center}
		\includegraphics[width=1\linewidth]{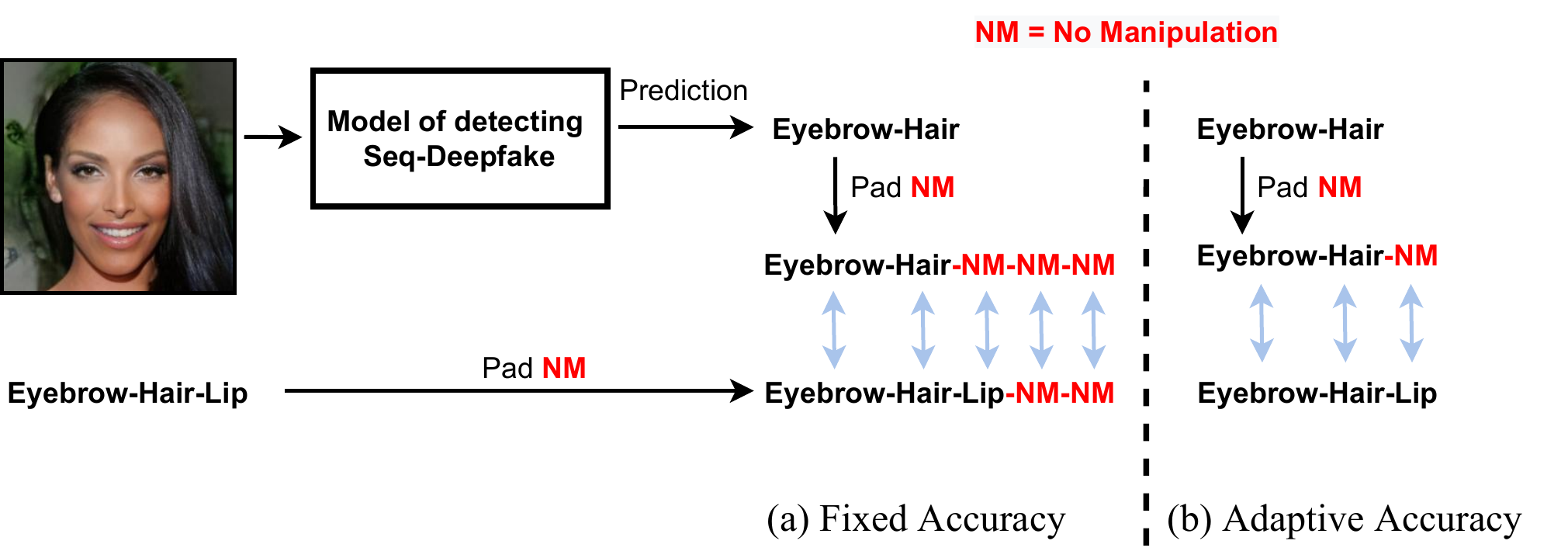}
	\end{center}
	\caption{Comparison between two evaluation metrics (a) Fixed Accuracy and (b) Adaptive Accuracy.}
	\label{fig:metric}
\end{figure}

As illustrated in Fig.~\ref{fig:metric}, we elaborate on two evaluation metrics proposed in the experiment for our new task.
\begin{itemize}
\item \textbf{Fixed Accuracy (Fixed-Acc):} As mentioned in the main paper, in the training process, since all of the baselines are only able to predict the facial manipulation with fixed length ($N=5$), `no manipulation' class will be padded into the annotation sequence if its length is shorter than $N$ so that we can keep the same length between predictions and annotation sequences for training. Following this strategy, as shown in Fig.~\ref{fig:metric}, under the evaluation metric of Fixed Accuracy, given the prediction, such as `Eyebrow-Hair', from the model of detecting Seq-Deepfake, we first pad `no manipulation' class into it to form the padded prediction sequence as `Eyebrow-Hair-NM-NM-NM' (NM means `no manipulation' class) so that we can obtain the prediction with fixed $N$-length ($N=5$). To keep the same length between predictions and annotation sequences for evaluation, we pad `no manipulation' class into the annotation of manipulation sequences as well, denoted as `Eyebrow-Hair-Lip-NM-NM', and compare each manipulation class in the predicted sequences with its corresponding annotation to calculate the evaluation accuracy.

\item \textbf{Adaptive Accuracy (Adaptive-Acc):} Moreover, since the proposed method exploits sequential information to detect facial manipulation sequences based on the auto-regressive mechanism, predictions will be automatically stopped once predicting the EOS token. Thus, the proposed method can detect facial manipulation sequences with adaptive lengths. To conduct the evaluation in this scenario, as illustrated in Fig.~\ref{fig:metric}, the second type of evaluation is devised, which compares predicted manipulations and corresponding annotations within the maximum steps of manipulations ($N = 3$ in Fig.~\ref{fig:metric} and we just pad one `no manipulation' class into prediction sequence) between them. This makes the evaluation focus more on accuracy of manipulations.
\end{itemize}

\section{Multi-head version of SECA}
\begin{figure}[t] 
	\begin{center}
		\includegraphics[width=1\linewidth]{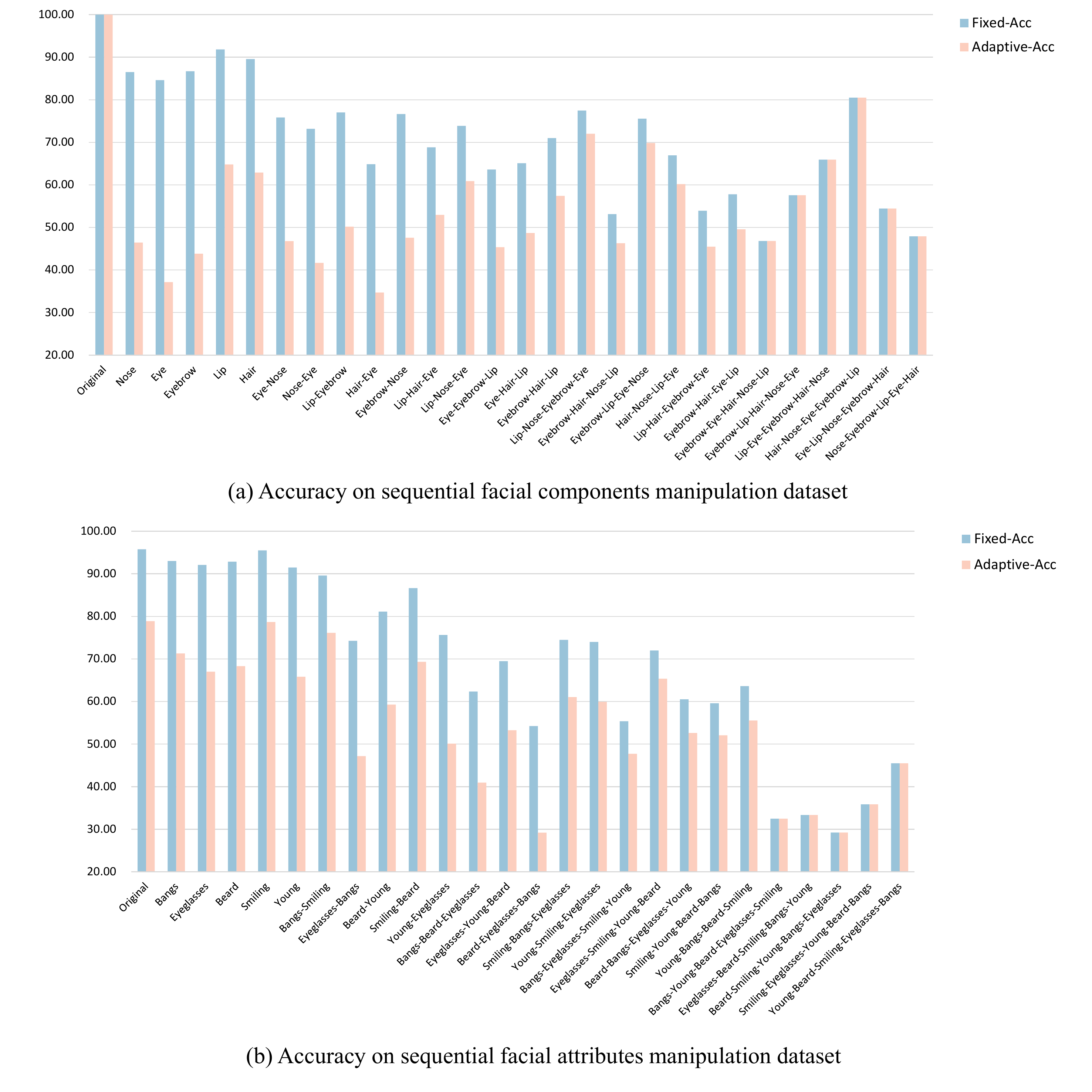}
	\end{center}
	\caption{Accuracy for each manipulation sequence.}
	\label{fig:acc}
\end{figure}

\begin{figure}[t] 
	\begin{center}
		\includegraphics[width=1\linewidth]{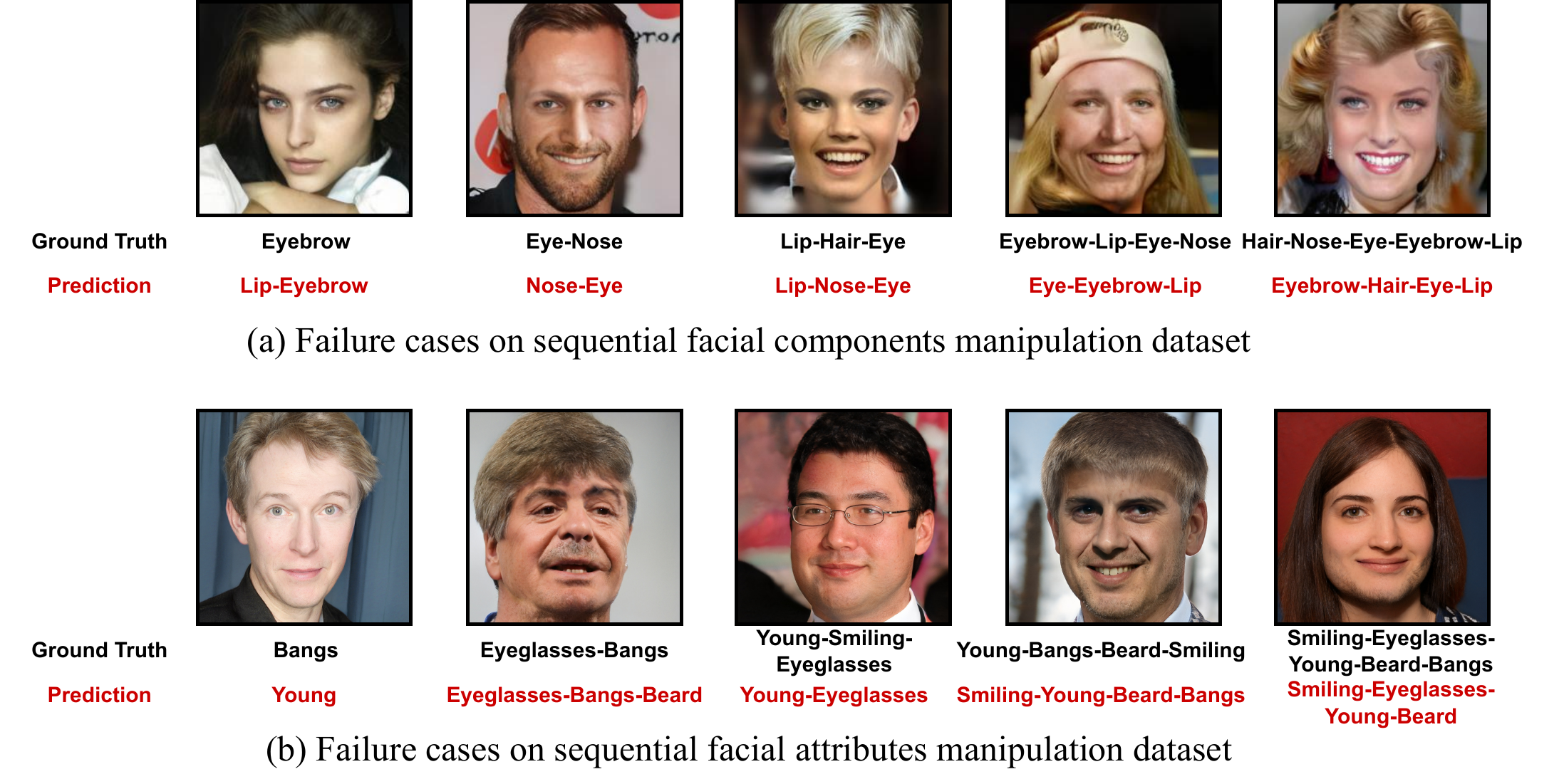}
	\end{center}
	\caption{Examples of failure cases.}
	\label{fig:failure}
\end{figure}

\begin{figure}[t]
	\begin{center}
		\includegraphics[height=11cm, width=1\linewidth]{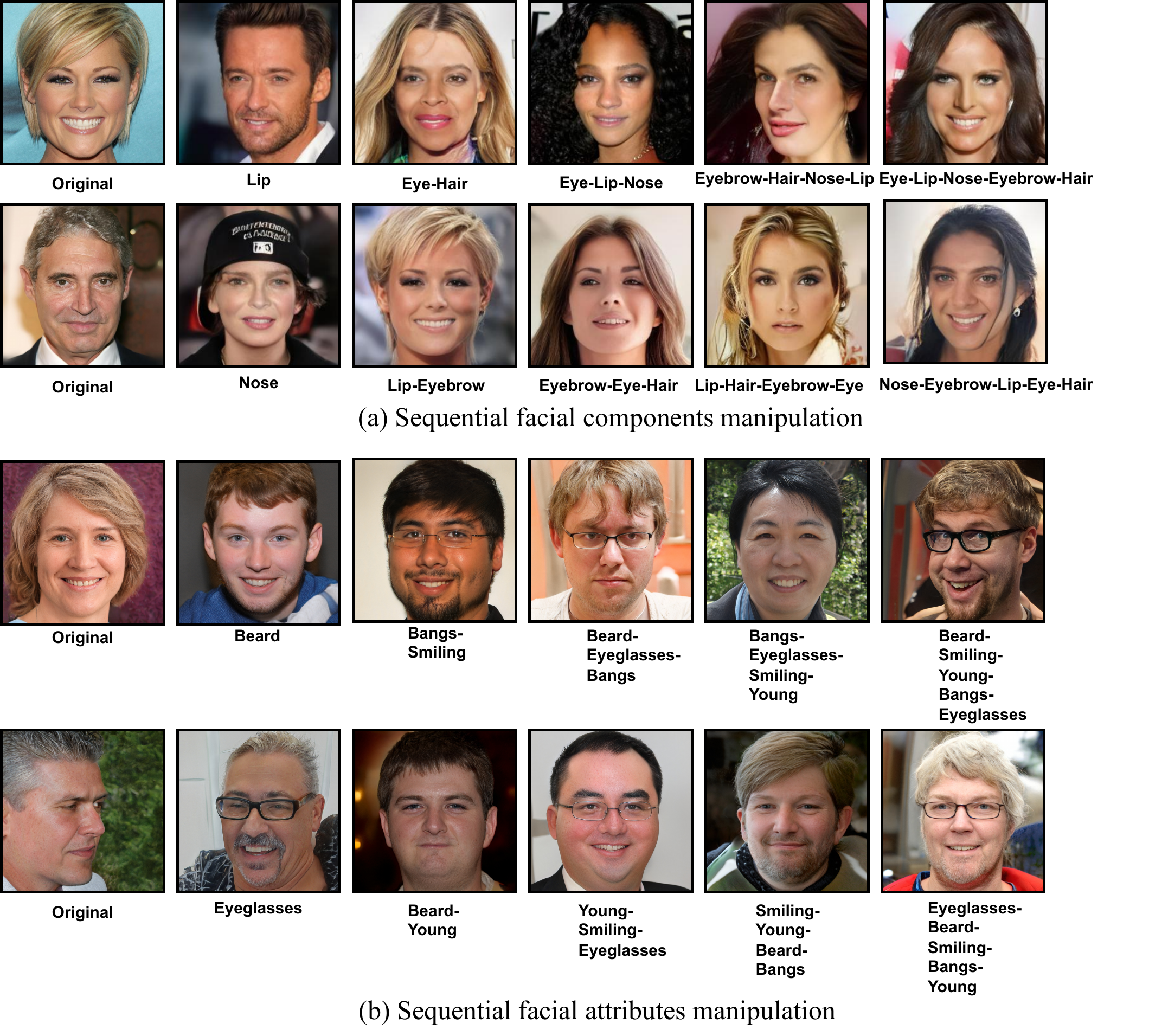}
	\end{center}
	\caption{Illustration of Seq-Deepfake dataset. Samples of Seq-Deepfake are provided with annotations of manipulation sequences.}
	\label{fig:DatasetExp-appendix}
\end{figure}

\begin{figure}[t]
	\begin{center}
		\includegraphics[height=9cm, width=1\linewidth]{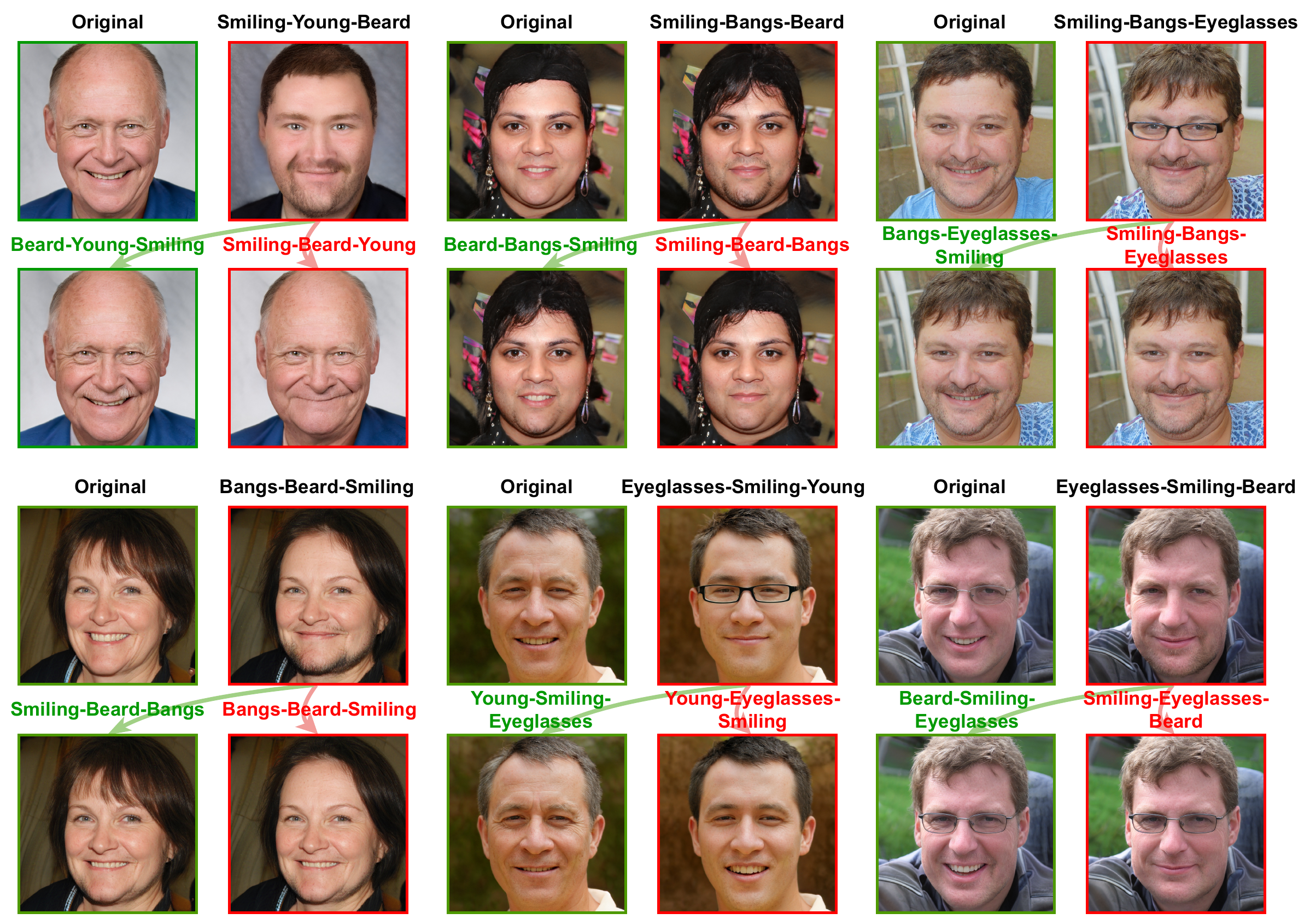}
	\end{center}
    \caption{Face recovery based on correct and wrong facial manipulation sequences.}
	\label{fig:facerecover-appendix}
\end{figure}

Similar to~\cite{gao2021fast}, we extend the basic version of Spatially Enhanced Cross-Attention (SECA) introduced in the main paper into multi-heads version, which enhances cross-attention features differently for different cross-attention heads. As mentioned in Eq.~\ref{equ:centerscale} in the main paper, the basic version of SECA estimates the 2-dimensional coordinates corresponding to spatial centers [$t_h, t_w$]. Similarly, the multi-head version of SECA estimates a head-shard spatial center [$t_h, t_w$] and then predicts a head-specific center offset [$\triangle t_{h,i}, \triangle t_{w,i}$] and corresponding head-specific scales [$r_{h,i}, r_{w,i}$] for $i$-th cross-attention head. In this way, we generate $i$-th head-specific Gaussian-shape spatial weight map $M_i$ based on the $i$-th head-specific center [$t_h + \triangle t_{h,i}, t_w + \triangle t_{w,i}$] and scales [$r_{h,i}, r_{w,i}$] as:
\begin{equation}
\begin{split}
& M_i(h, w) =  \text{exp}\left(-\frac{(h-(t_h + \triangle t_{h,i}))^2}{\lambda r_{h,i}^2}-\frac{(w-(t_w + \triangle t_{w,i}))^2}{\lambda r_{w,i}^2} \right)
\end{split}
\label{equ:spatialmap-appendix}
\end{equation}
\noindent Based on this, we can calculate the features of sequential relation $f^{seq}_i$ from $i$-th cross-attention head enhanced by the $i$-th SECA as follows:
\begin{equation}
\begin{split}
& f^{seq}_i = \text{Softmax}(K_i^T Q_i \sqrt{d} + logM_i)V_i, \\
\end{split}
\label{equ:decoder-appendix}
\end{equation}
\noindent Different from basic version of SECA, above Eq.~\ref{equ:decoder-appendix} shows that in the multi-head version of SECA, the cross-attention of the $i$-th head is element-wise added with logarithm of $i$-th head-specific spatial weight map $M_i$, which contributes to a more adaptive and specific enhanced cross-attention. Experiments regarding the proposed method in the main paper are all carried out based on the multi-head version of SECA.

\section{Accuracy For Each Manipulation Sequence}

As mentioned in the main paper, we generate 28 types of manipulation sequences based on facial components manipulation while 26 types of manipulation sequences based on facial attributes manipulation. To provide a more detailed analysis, in this section, we plot accuracy for each manipulation sequence in both facial manipulation methods as shown in Fig.~\ref{fig:acc}. It can be observed that diverse accuracy performance are achieved for different manipulation sequences, ranging from 46.81\% to 100\% under Fixed-Acc and 34.69\% to 100\% under Adaptive-Acc in sequential facial components manipulation, while ranging from 29.25\% to 95.75\% under Fixed-Acc and 29.21\% to 78.88\% under Adaptive-Acc in sequential facial attributes manipulation. This demonstrates various manipulation sequences are challenging for detection and there exist some extremely hard cases. Therefore, we should further improve our method to cope with all types of manipulation sequences in the future. Furthermore, it can be seen from Fig.~\ref{fig:acc} that the accuracy gap between two evaluation metrics, Fixed-Acc and Adaptive-Acc, decreases along with the length of sequence increases. This is because the padded `no manipulation' class is fewer in the longer manipulation sequence when evaluating under Adaptive-Acc, which is closer to the evaluation under Fixed-Acc.

\section{Failure Cases}

To provide a deeper understanding for our novel task and method, we display some failure cases produced by the proposed method as illustrated in Fig.~\ref{fig:failure}. From Fig.~\ref{fig:failure}, it can be seen that there exist diverse failure cases, including wrong predictions with respect to manipulation type, sequence order, sequence length, etc. This validates that it is quite difficult for our novel research problem since we need to detect facial manipulation sequences in terms of correct manipulation types, orders and lengths simultaneously from hyper-realistic face images with subtle sequential manipulations. This motivates us to continually improve the performance of the proposed SeqFakeFormer to tackle such a challenging task in the future.

\section{Sequential Deepfake Dataset}

We display more samples from the generated large-scale Sequential Deepfake (Seq-Deepfake) dataset in Fig.~\ref{fig:DatasetExp-appendix}. As shown in Fig.~\ref{fig:DatasetExp-appendix}, based on two different facial manipulation methods, facial components manipulation~\cite{kim2021exploiting} and facial attributes manipulation~\cite{jiang2021talk}, various sequential facial manipulations are produced with diverse manipulation steps, expressions, ages, and genders.

\section{Face Recovery}

Fig.~\ref{fig:facerecover-appendix} shows more samples regarding the face recovery based on correct and wrong facial manipulation sequences. As can be observed in Fig.~\ref{fig:facerecover-appendix}, once we detect the correct facial manipulation sequence, \textit{i.e.} correct manipulations ordered with correct manipulation steps, we can recover original face by performing face attribute manipulation based on the inverse order of detected facial manipulation sequence (process with green arrow). In contrast, recovering the face image with wrongly ordered manipulation sequences may encounter different problems, such as incomplete recovery of age, smile, glasses, bangs, etc. (process with red arrow).

\end{document}